\documentclass{article}

\PassOptionsToPackage{numbers, compress}{natbib}

\usepackage[final]{nips_2017}

\usepackage[utf8]{inputenc} 
\usepackage[T1]{fontenc}    
\usepackage{hyperref}       
\usepackage{url}            
\usepackage{booktabs}       
\usepackage{amsfonts}       
\usepackage{nicefrac}       
\usepackage{microtype}      

\usepackage{amsmath}

\usepackage{bm}

\def\abovestrut#1{\rule[0in]{0in}{#1}\ignorespaces}
\def\belowstrut#1{\rule[-#1]{0in}{#1}\ignorespaces}

\def\abovespace{\abovestrut{0.20in}}

\def\belowspace{\belowstrut{0.10in}}

\usepackage{algorithm}
\usepackage{algorithmic}
\renewcommand{\algorithmiccomment}[1]{\hfill$\triangleright$ #1}

\usepackage{graphicx} 
\usepackage{caption}
\usepackage{subcaption}
\usepackage{wrapfig}

\title{Prototypical Networks for Few-shot Learning}

\author{
  Jake Snell  \\ University of Toronto\footnote{Initial work done while at Twitter.} \And
  Kevin Swersky \\ Twitter \And
  Richard S. Zemel \\ University of Toronto, Vector Institute
}

\begin{document}

\maketitle

\begin{abstract} 
We propose \emph{prototypical networks} for the problem of few-shot classification, where a classifier must generalize
to new classes not seen in the training set, given only a small number of
examples of each new class. Prototypical networks
learn a metric space in which classification can be performed by computing 
distances to prototype representations of each class. Compared to recent
approaches for few-shot learning, they reflect a simpler
inductive bias that is beneficial in this limited-data regime, and achieve
excellent results. We provide an analysis showing
that some simple design decisions can yield substantial improvements
over recent approaches involving complicated architectural choices and
meta-learning. We further extend prototypical networks
to zero-shot learning and achieve state-of-the-art results
on the CU-Birds dataset.
\end{abstract} 

\section{Introduction}

Few-shot classification \citep{miller2000learning,lake2011one,koch2015siamese} is a task in which a classifier must be adapted to accommodate new classes not seen in training, given only a few examples of each of these classes. A naive approach, such as re-training the model on the new data, would severely overfit. While the problem is quite difficult, it has been demonstrated that humans have the ability to perform even one-shot classification, where only a single example of each new class is given, with a high degree of accuracy~\citep{lake2011one}. 

Two recent approaches have made significant progress in few-shot learning.
\citet{vinyals2016matching} proposed \emph{matching networks}, which uses an attention mechanism over a learned embedding of the labeled set of examples (the \emph{support set}) to predict classes for the unlabeled points (the \emph{query set}). Matching networks can be interpreted as a weighted nearest-neighbor classifier applied within an embedding space. Notably, this model utilizes sampled mini-batches called \emph{episodes} during training, where each episode is designed to mimic the few-shot task by subsampling classes as well as data points. The use of episodes makes the training problem more faithful to the test environment and thereby improves generalization.
\citet{ravi2017meta} take the episodic training idea further and propose a meta-learning approach to few-shot learning. Their approach involves training an LSTM~\cite{hochreiter1997long} to produce the updates to a classifier, given an episode, such that it will generalize well to a test-set. Here, rather than training a single model over multiple episodes, the LSTM meta-learner learns to train a custom model for each episode.

We attack the problem of few-shot learning by addressing the key issue of overfitting. Since data is severely limited, we work under the assumption that a classifier should have a very simple inductive bias. Our approach, {\it prototypical networks}, is based on the idea that there exists an embedding in which points cluster around a single prototype representation for each class. In order to do this, we learn a non-linear mapping of the input into an embedding space using a neural network and take a class's prototype to be the mean of its support set in the embedding space. Classification is then performed for an embedded query point by simply finding the nearest class prototype. We follow the same approach to tackle zero-shot learning; here each class comes with meta-data giving a high-level description of the class rather than a small number of labeled examples. We therefore learn an embedding of the meta-data into a shared space to serve as the prototype for each class. Classification is performed, as in the few-shot scenario, by finding the nearest class prototype for an embedded query point.

In this paper, we formulate prototypical networks for both the few-shot and zero-shot settings. We draw connections to matching networks in the one-shot setting, and analyze the underlying distance function used in the model. In particular, we relate prototypical networks to clustering \cite{banerjee2005clustering} in order to justify the use of class means as prototypes when distances are computed with a Bregman divergence, such as squared Euclidean distance. We find empirically that the choice of distance is vital, as Euclidean distance greatly outperforms the more commonly used cosine similarity. On several benchmark tasks, we achieve state-of-the-art performance. Prototypical networks are simpler and more efficient than recent meta-learning algorithms, making them an appealing approach to few-shot and zero-shot learning.
\begin{figure}[tb]
    \centering
    \begin{subfigure}{.5\textwidth}
        \centering
        \includegraphics[height=1.4in]{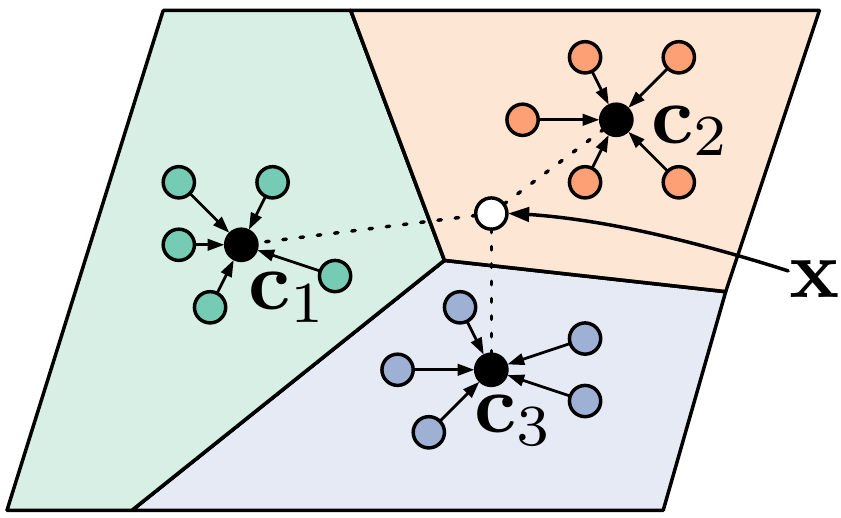}
        \caption{Few-shot}
    \end{subfigure}\hfill%
    \begin{subfigure}{.5\textwidth}
        \centering
        \includegraphics[height=1.4in]{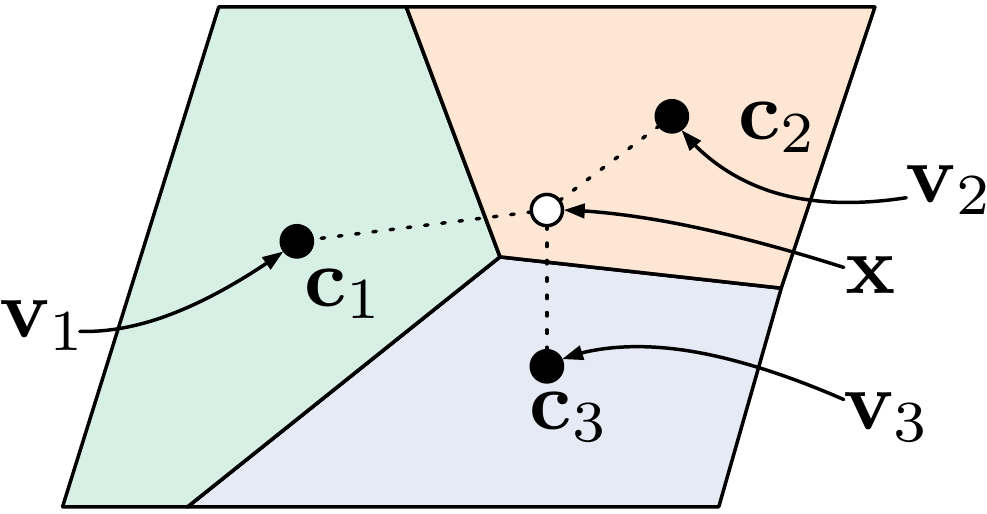}
        \caption{Zero-shot}
    \end{subfigure}
    \caption{Prototypical networks in the few-shot and zero-shot scenarios.
    \textbf{Left}: Few-shot prototypes $\mathbf{c}_k$ are computed as the mean of embedded support examples for each class. \textbf{Right}: Zero-shot prototypes $\mathbf{c}_k$ are produced by embedding class meta-data $\mathbf{v}_k$. In either case, embedded query points are classified via a softmax over distances to class prototypes: $p_{\bm \phi}(y=k|\mathbf{x}) \propto \exp(-d(f_{\bm \phi}(\mathbf{x}), \mathbf{c}_k))$.}
    \label{fig:protodiagram}
\end{figure}
\section{Prototypical Networks}
\subsection{Notation}
In few-shot classification we are given a small support set of $N$ labeled examples $S = \{ (\mathbf{x}_1, y_1), \ldots, (\mathbf{x}_N, y_N) \}$ where each $\mathbf{x}_i \in \mathbb{R}^D$ is the $D$-dimensional feature vector of an example and $y_i \in \{1, \ldots, K\}$ is the corresponding label. $S_k$ denotes the set of examples labeled with class $k$.

\subsection{Model}
Prototypical networks compute an $M$-dimensional representation $\mathbf{c}_k \in \mathbb{R}^M$, or \emph{prototype}, of each class through an embedding function $f_{\bm \phi} : \mathbb{R}^D \rightarrow \mathbb{R}^M$ with  learnable parameters $\bm{\phi}$. Each prototype is the mean vector of the embedded support points belonging to its class:
\begin{equation}
    \mathbf{c}_k = \frac{1}{|S_k|} \sum_{(\mathbf{x}_i, y_i) \in S_k} f_{\bm{\phi}}(\mathbf{x}_i)
   \label{eq:prototype}
\end{equation}
Given a distance function $d: \mathbb{R}^M \times \mathbb{R}^M \rightarrow [0, +\infty)$, prototypical networks produce a distribution over classes for a query point $\mathbf{x}$ based on a softmax over distances to the prototypes in the embedding space:
\begin{equation}
    p_{\bm \phi}(y = k\,|\,\mathbf{x}) = \frac{\exp(-d(f_{\bm \phi}(\mathbf{x}), \mathbf{c}_k))}{\sum_{k'} \exp(-d(f_{\bm \phi}(\mathbf{x}), \mathbf{c}_{k'}))}
    \label{eq:classdist}
\end{equation}
Learning proceeds by minimizing the negative log-probability $J(\bm{\phi}) = -\log p_{\bm \phi}(y = k \,|\,\mathbf{x})$ of the true class $k$ via SGD. Training episodes are formed by randomly selecting a subset of classes from the training set, then choosing a subset of examples within each class to act as the support set and a subset of the remainder to serve as query points. Pseudocode to compute the loss $J(\bm{\phi})$ for a training episode is provided in Algorithm~\ref{alg:prototrain}.

\begin{algorithm*}[tb]
    \caption{Training episode loss computation for prototypical networks. $N$ is the number of examples in the training set, $K$ is the number of classes in the training set, $N_C \le K$ is the number of classes per episode, $N_S$ is the number of support examples per class, $N_Q$ is the number of query examples per class. $\textsc{RandomSample}(S, N)$ denotes a set of $N$ elements chosen uniformly at random from set $S$, without replacement.}
    \label{alg:prototrain}
    \begin{algorithmic}
    \REQUIRE Training set $\mathcal{D} = \{ (\mathbf{x}_1, y_1), \ldots, (\mathbf{x}_N, y_N) \}$, where each $y_i \in \{1, \ldots, K\}$. $\mathcal{D}_k$ denotes the subset of $\mathcal{D}$ containing all elements $(\mathbf{x}_i, y_i)$ 
    such that $y_i = k$. 
    \ENSURE The loss $J$ for a randomly generated training episode.
    \STATE $V \gets \textsc{RandomSample}(\{1, \ldots, K\}, N_C)$ 
        \COMMENT{Select class indices for episode}
    \FOR{$k$ in $\{1, \ldots, N_C\}$}
        \STATE $S_k \gets \textsc{RandomSample}(\mathcal{D}_{V_k}, N_S)$ \COMMENT{Select support examples}
        \STATE $Q_k \gets \textsc{RandomSample}(\mathcal{D}_{V_k}
            \setminus S_k, N_Q)$ \COMMENT{Select query examples}
        \STATE $\displaystyle \mathbf{c}_k \gets \frac{1}{N_C} \sum_{(\mathbf{x}_i, y_i) \in S_k} f_{\bm \phi}(\mathbf{x}_i)$ \COMMENT{Compute prototype from support examples}
    \ENDFOR
    \STATE $J \gets 0$ \COMMENT{Initialize loss}
    \FOR{$k$ in $\{1, \ldots, N_C\}$}
        \FOR{$(\mathbf{x}, y)$ in $Q_k$}
            \STATE $\displaystyle J \gets J + 
            \frac{1}{N_C N_Q} \left[
                d(f_{\bm \phi}(\mathbf{x}), \mathbf{c}_k)) + \log \sum_{k'} \exp(-d(f_{\bm \phi}(\mathbf{x}), \mathbf{c}_k))
            \right]$ \COMMENT{Update loss}
        \ENDFOR
    \ENDFOR
    \end{algorithmic}
\end{algorithm*}

\subsection{Prototypical Networks as Mixture Density Estimation}
\label{sec:bregman}

For a particular class of distance functions, known as \emph{regular Bregman divergences} \cite{banerjee2005clustering}, the prototypical networks algorithm is equivalent to performing mixture density estimation on the support set with an exponential family density. A regular Bregman divergence $d_\varphi$ is defined as:
\begin{equation}
    d_{\varphi}(\mathbf{z}, \mathbf{z}') = \varphi(\mathbf{z}) - \varphi(\mathbf{z}') - (\mathbf{z} - \mathbf{z}')^T \nabla \varphi(\mathbf{z}'),
\end{equation}
where $\varphi$ is a differentiable, strictly convex function of the Legendre type. Examples of Bregman divergences include squared Euclidean distance $\|\mathbf{z} - \mathbf{z}'\|^2$ and Mahalanobis distance.

Prototype computation can be viewed in terms of hard clustering on the support set, with one cluster per class and each support point assigned to its corresponding class cluster. It has been shown \cite{banerjee2005clustering} for Bregman divergences that the cluster representative achieving minimal distance to its assigned points is the cluster mean. Thus the prototype computation in Equation \eqref{eq:prototype} yields optimal cluster representatives given the support set labels when a Bregman divergence is used.

Moreover, any regular exponential family distribution $p_\psi(\mathbf{z} | \bm{\theta})$ with parameters $\bm{\theta}$ and cumulant function $\psi$ can be written in terms of a uniquely determined regular Bregman divergence \cite{banerjee2005clustering}:
\begin{equation}
     p_\psi(\mathbf{z} | \bm{\theta}) = \exp \{ \mathbf{z}^T \bm{\theta} - \psi(\bm{\theta}) - g_\psi(\mathbf{z}) \}
     = \exp \{ -d_{\varphi}(\mathbf{z}, \bm{\mu}(\bm{\theta})) - g_\varphi(\mathbf{z}) \}
\end{equation}
Consider now a regular exponential family mixture model with parameters $\bm{\Gamma} = \{\bm{\theta}_k, \pi_k\}_{k=1}^K$:
\begin{equation}
     p(\mathbf{z} | \bm{\Gamma}) = \sum_{k=1}^K \pi_k p_\psi(\mathbf{z} | \bm{\theta}_k) = \sum_{k=1}^K \pi_k \exp(-d_{\varphi}(\mathbf{z}, \bm{\mu}(\bm{\theta}_k)) - g_\varphi(\mathbf{z}))
\end{equation}
Given $\bm{\Gamma}$, inference of the cluster assignment $y$ for an unlabeled point $\mathbf{z}$ becomes:
\begin{equation}
    p(y = k | \mathbf{z}) = \frac{\pi_k \exp(-d_{\varphi}(\mathbf{z}, \bm{\mu}(\bm{\theta}_k)))}
                                 {\sum_{k'} \pi_{k'} \exp(-d_{\varphi}(\mathbf{z}, \bm{\mu}(\bm{\theta}_k)))}
    \label{eq:clusterinf}
\end{equation}
For an equally-weighted mixture model with one cluster per class, cluster assignment inference \eqref{eq:clusterinf} is equivalent to query class prediction \eqref{eq:classdist} with $f_{\phi}(\mathbf{x}) = \mathbf{z}$ and $\mathbf{c}_k = \bm{\mu}(\bm{\theta}_k)$.
In this case, prototypical networks are effectively performing mixture density estimation with an exponential family distribution determined by $d_\varphi$. The choice of distance therefore specifies modeling assumptions about the class-conditional data distribution in the embedding space.
\subsection{Reinterpretation as a Linear Model}

A simple analysis is useful in gaining insight into the nature of the learned classifier.
When we use Euclidean distance $d(\mathbf{z}, \mathbf{z'}) = \|\mathbf{z} - \mathbf{z}'\|^2$, then the model in Equation \eqref{eq:classdist} is equivalent to a linear model with a particular parameterization~\cite{mensink2013distance}. To see this, expand the term in the exponent:
\begin{align}
    -\| f_{\bm \phi}(\mathbf{x}) - \mathbf{c}_k \|^2 &=  -f_{\bm \phi}(\mathbf{x})^\top f_{\bm \phi}(\mathbf{x}) + 2\mathbf{c}_k^\top f_{\bm \phi}(\mathbf{x}) - \mathbf{c}_k^\top \mathbf{c}_k \label{eq:expanded}
\end{align}
The first term in Equation \eqref{eq:expanded} is constant with respect to the class $k$, so it does not affect the softmax probabilities. We can write the remaining terms as a linear model as follows:
\begin{equation}
     2\mathbf{c}_k^\top f_{\bm \phi}(\mathbf{x}) - \mathbf{c}_k^\top \mathbf{c}_k = \mathbf{w}_k^\top f_{\bm \phi}(\mathbf{x}) + b_k \mbox{, where  }
     \mathbf{w}_k = 2\mathbf{c}_k \mbox{ and }
     b_k = -\mathbf{c}_k^\top \mathbf{c}_k
\end{equation}
We focus primarily on squared Euclidean distance (corresponding to spherical Gaussian densities) in this work. Our results indicate that Euclidean distance is an effective choice despite the equivalence to a linear model. We hypothesize this is because all of the required non-linearity can be learned within the embedding function.
Indeed, this is the approach that modern neural network classification systems currently use, e.g., \citep{krizhevsky2012imagenet, szegedy2015going}.

\subsection{Comparison to Matching Networks}
\label{sec:matchingcomparison}

Prototypical networks differ from matching networks in the few-shot case with equivalence in the one-shot scenario. Matching networks \cite{vinyals2016matching} produce a weighted nearest neighbor classifier given the support set, while prototypical networks produce a linear classifier when squared Euclidean distance is used. In the case of one-shot learning, $\mathbf{c}_k=\mathbf{x}_k$ since there is only one support point per class, and matching networks and prototypical networks become equivalent.

A natural question is whether it makes sense to use multiple prototypes per class instead of just one.
If the number of prototypes per class is fixed and greater than $1$, then this would require a partitioning scheme to further cluster the support points within a class. This has been proposed in \citet{mensink2013distance} and \citet{rippel2015metric}; however both methods require a separate partitioning phase that is decoupled from the weight updates, while our approach is simple to learn with ordinary gradient descent methods.

\citet{vinyals2016matching} propose a number of extensions, including decoupling the embedding functions of the support and query points, and using a second-level, fully-conditional embedding (FCE) that takes into account specific points in each episode. These could likewise be incorporated into prototypical networks, however they increase the number of learnable parameters, and FCE imposes an arbitrary ordering on the support set using a bi-directional LSTM. Instead, we show that it is possible to achieve the same level of performance using simple design choices, which we outline next.

\subsection{Design Choices}

\paragraph{Distance metric}
\citet{vinyals2016matching} and \citet{ravi2017meta} apply matching networks using cosine distance.
However for both prototypical and matching networks any distance is permissible, and we found that using squared Euclidean distance can greatly improve results for both. We conjecture this is primarily due to cosine distance not being a Bregman divergence, and thus the equivalence to mixture density estimation discussed in Section~\ref{sec:bregman} does not hold.

\paragraph{Episode composition}
A straightforward way to construct episodes, used in \citet{vinyals2016matching} and \citet{ravi2017meta}, is to choose $N_c$ classes and $N_S$ support points per class in order to match the expected situation at test-time. That is, if we expect at test-time to perform $5$-way classification and $1$-shot learning, then training episodes could be comprised of $N_c=5$, $N_S=1$. We have found, however, that it can be extremely beneficial to train with a higher $N_c$, or ``way'', than will be used at test-time. In our experiments, we tune the training $N_c$ on a held-out validation set. Another consideration is whether to match $N_S$, or ``shot'', at train and test-time.
For prototypical networks, we found that it is usually best to train and test with the same ``shot'' number.

\subsection{Zero-Shot Learning}
Zero-shot learning differs from few-shot learning in that instead of being given a support set of training points, we are given a class meta-data vector $\mathbf{v}_k$ for each class. These could be determined in advance, or they could be learned from e.g., raw text~\citep{elhoseiny2013write}. Modifying prototypical networks to deal with the zero-shot case is straightforward: we simply define $\mathbf{c}_k=g_{\bm \vartheta}(\mathbf{v}_k)$ to be a separate embedding of the meta-data vector. An illustration of the zero-shot procedure for prototypical networks as it relates to the few-shot procedure is shown in Figure~\ref{fig:protodiagram}. Since the meta-data vector and query point come from different input domains, we found it was helpful empirically to fix the prototype embedding $g$ to have unit length, however we do not constrain the query embedding $f$.

\section{Experiments}

For few-shot learning, we performed experiments on Omniglot \citep{lake2011one} and the \emph{mini}ImageNet version of ILSVRC-2012 \citep{russakovsky2015imagenet} with the splits proposed by \citet{ravi2017meta}.  We perform zero-shot experiments on the 2011 version of the Caltech UCSD bird dataset (CUB-200 2011) \citep{welinder2010caltech}.

\subsection{Omniglot Few-shot Classification}
Omniglot \citep{lake2011one} is a dataset of 1623 handwritten characters collected from 50 alphabets. There are 20 examples associated with each character, where each example is drawn by a different human subject. We follow the procedure of \citet{vinyals2016matching} by resizing the grayscale images to 28 $\times$ 28 and augmenting the character classes with rotations in multiples of 90 degrees. We use 1200 characters plus rotations for training (4,800 classes in total) and the remaining classes, including rotations, for test. Our embedding architecture mirrors that used by \citet{vinyals2016matching} and is composed of four convolutional blocks. Each block comprises a 64-filter 3 $\times$ 3 convolution, batch normalization layer \citep{ioffe2015batch}, a ReLU nonlinearity and a 2 $\times$ 2 max-pooling layer. When applied to the 28 $\times$ 28 Omniglot images this architecture results in a 64-dimensional output space. We use the same encoder for embedding both support and query points. All of our models were trained via SGD with Adam \citep{kingma2014adam}. We used an initial learning rate of $10^{-3}$ and cut the learning rate in half every 2000 episodes. No regularization was used other than batch normalization.

We trained prototypical networks using Euclidean distance in the 1-shot and 5-shot scenarios with training episodes containing 60 classes and 5 query points per class. We found that it is advantageous to match the training-shot with the test-shot, and to use more classes (higher ``way'') per training episode rather than fewer. We compare against various baselines, including the neural statistician \citep{edwards2017towards} and both the fine-tuned and non-fine-tuned versions of matching networks \citep{vinyals2016matching}. We computed classification accuracy for our models averaged over 1000 randomly generated episodes from the test set. The results are shown in Table~\ref{tab:omniglot} and to our knowledge they represent the state-of-the-art on this dataset.

\begin{table*}[tb]
\caption{Few-shot classification accuracies on Omniglot.}
\vskip -0.1in
\label{tab:omniglot}
\begin{center}
\begin{small}
\begin{tabular}{lcccccc}
\hline
\abovespace
& & & \multicolumn{2}{c}{\textbf{5-way Acc.}} & \multicolumn{2}{c}{\textbf{20-way Acc.}} \\
\belowspace
\textbf{Model} & \textbf{Dist.} & \textbf{Fine Tune} & 1-shot & 5-shot & 1-shot & 5-shot \\
\hline
\abovespace
\textbf{\textsc{Matching Networks}} \citep{vinyals2016matching} & Cosine & N & 98.1\% & 98.9\% & 93.8\% & 98.5\% \\
\textbf{\textsc{Matching Networks}} \citep{vinyals2016matching} & Cosine & Y & 97.9\% & 98.7\% & 93.5\% & 98.7\% \\
\textbf{\textsc{Neural Statistician}} \citep{edwards2017towards} & - & N & 98.1\% & 99.5\% & 93.2\% & 98.1\% \\
\belowspace
\textbf{\textsc{Prototypical Networks (Ours)}} & Euclid. & N & \textbf{98.8\%} & \textbf{99.7\%} & \textbf{96.0\%} & \textbf{98.9\%} \\
\hline
\end{tabular}
\end{small}
\end{center}
\vskip -0.1in
\end{table*}

\subsection{\emph{mini}ImageNet Few-shot Classification}
\label{sec:miniimagenet}
The \emph{mini}ImageNet dataset, originally proposed by \citet{vinyals2016matching}, is derived from the larger ILSVRC-12 dataset~\citep{russakovsky2015imagenet}. The splits used by \citet{vinyals2016matching} consist of 60,000 color images of size 84 $\times$ 84 divided into 100 classes with 600 examples each. For our experiments, we use the splits introduced by \citet{ravi2017meta} in order to directly compare with state-of-the-art algorithms for few-shot learning. Their splits use a different set of 100 classes, divided into 64 training, 16 validation, and 20 test classes. We follow their procedure by training on the 64 training classes and using the 16 validation classes for monitoring generalization performance only.

We use the same four-block embedding architecture as in our Omniglot experiments, though here it results in a 1600-dimensional output space due to the increased size of the images. We also use the same learning rate schedule as in our Omniglot experiments and train until validation loss stops improving. We train using 30-way episodes for 1-shot classification and 20-way episodes for 5-shot classification. We match train shot to test shot and each class contains 15 query points per episode.  We compare to the baselines as reported by \citet{ravi2017meta}, which include a simple nearest neighbor approach on top of features learned by a classification network on the 64 training classes. The other baselines are two non-fine-tuned variants of matching networks (both ordinary and FCE) and the Meta-Learner LSTM.
As can be seen in Table~\ref{tab:miniImagenet}, prototypical networks achieves state-of-the-art here by a wide margin.
\begin{table*}[tb]
\caption{Few-shot classification accuracies on \textit{mini}ImageNet. All accuracy results are averaged over 600 test episodes and are reported with 95\% confidence intervals. \textsuperscript{$\ast$}Results reported by \cite{ravi2017meta}.}
\vskip -0.1in
\label{tab:miniImagenet}
\begin{center}
\begin{small}
\begin{tabular}{lcccc}
\hline
\abovespace
&&& \multicolumn{2}{c}{\textbf{5-way Acc.}} \\
\belowspace
\textbf{Model} & \textbf{Dist.} & \textbf{Fine Tune} & 1-shot & 5-shot \\
\hline
\abovespace
\textbf{\textsc{Baseline Nearest neighbors}}\textsuperscript{$\ast$} & Cosine & N & 28.86 $\pm$ 0.54\% & 49.79 $\pm$ 0.79\% \\
\textbf{\textsc{Matching Networks}} \citep{vinyals2016matching}\textsuperscript{$\ast$} & Cosine & N & 43.40 $\pm$ 0.78\% & 51.09 $\pm$ 0.71\% \\
\textbf{\textsc{Matching Networks FCE}} \citep{vinyals2016matching}\textsuperscript{$\ast$} & Cosine & N & 43.56 $\pm$ 0.84\% & 55.31 $\pm$ 0.73\% \\
\textbf{\textsc{Meta-Learner LSTM}} \citep{ravi2017meta}\textsuperscript{$\ast$} & - & N & 43.44 $\pm$ 0.77\% & 60.60 $\pm$ 0.71\% \\
\belowspace
\textbf{\textsc{Prototypical Networks (Ours)}} & Euclid. & N & \textbf{49.42 $\pm$ 0.78\%} & \textbf{68.20 $\pm$ 0.66\%}\\
\hline
\end{tabular}
\end{small}
\end{center}
\end{table*}

\begin{figure*}[bt]
    \centering
    \includegraphics[width=0.9\textwidth]{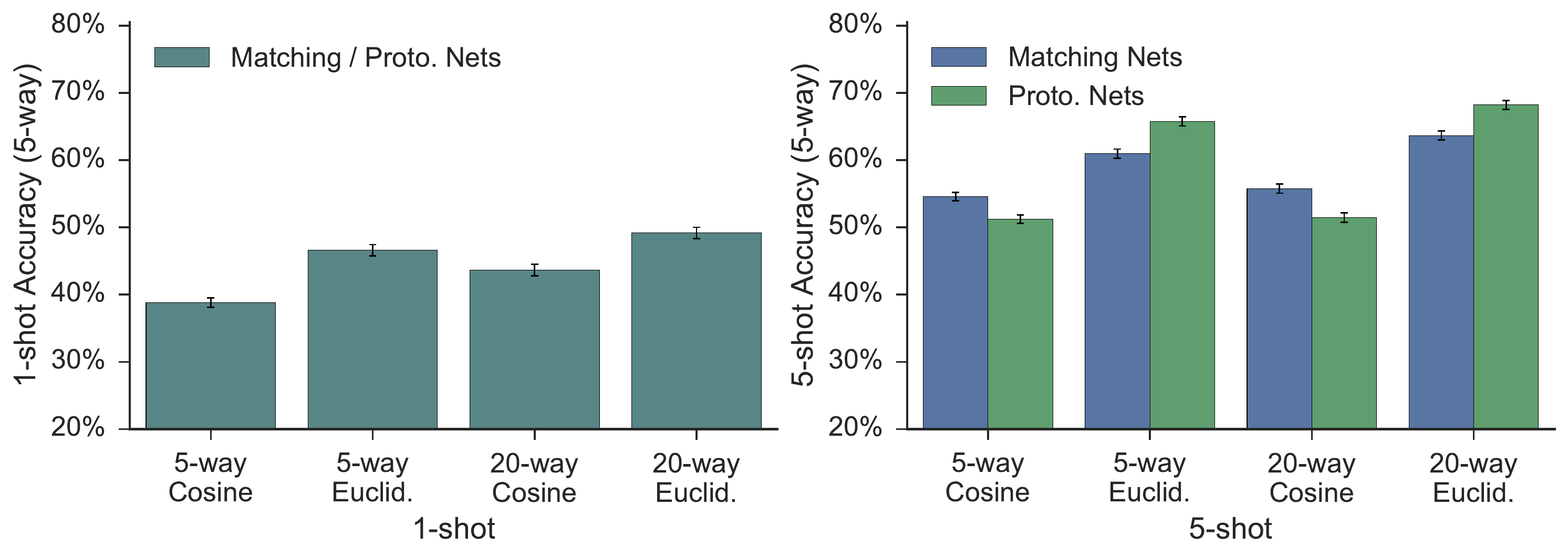}
    \caption{Comparison showing the effect of distance metric and number of classes per training episode on 5-way classification accuracy for both matching and prototypical networks on \emph{mini}ImageNet. The $x$-axis indicates
    configuration of the training episodes (way, distance, and shot), and the $y$-axis indicates 5-way test accuracy
    for the corresponding shot. Error bars indicate 95\% confidence intervals as computed over 600 test episodes. Note that matching networks and prototypical networks are identical in the 1-shot case.}
    \label{fig:config_analysis}
\end{figure*}

We conducted further analysis, to determine the effect of distance metric and the number of training classes per episode on the performance of prototypical networks and matching networks. To make the methods comparable, we use our own implementation of matching networks that utilizes the same embedding architecture as our prototypical networks. 
In Figure~\ref{fig:config_analysis} we compare cosine vs. Euclidean distance and 5-way vs. 20-way training episodes in the 1-shot and 5-shot scenarios, with 15 query points per class per episode. We note that 20-way achieves higher accuracy than 5-way and conjecture that the increased difficulty of 20-way classification helps the network to generalize better, because it forces the model to make more fine-grained decisions in the embedding space. Also, using Euclidean distance improves performance substantially over cosine distance. This effect is even more pronounced for prototypical networks, in which computing the class prototype as the mean of embedded support points is more naturally suited to Euclidean distances since cosine distance is not a Bregman divergence.

\subsection{CUB Zero-shot Classification}

In order to assess the suitability of our approach for zero-shot learning, we also run experiments on the Caltech-UCSD Birds (CUB) 200-2011 dataset \citep{welinder2010caltech}. The CUB dataset contains 11,788 images of 200 bird species. We closely follow the procedure of \citet{reed2016learning} in preparing the data. We use their splits to divide the classes into 100 training, 50 validation, and 50 test. For images we use 1,024-dimensional features extracted by applying GoogLeNet \citep{szegedy2015going} to middle, upper left, upper right, lower left, and lower right crops of the original and horizontally-flipped image\footnote{Features downloaded from \url{https://github.com/reedscot/cvpr2016}.}. At test time we use only the middle crop of the original image. For class meta-data we use the 312-dimensional continuous attribute vectors provided with the CUB dataset. These attributes encode various characteristics of the bird species such as their color, shape, and feather patterns.

We learned a simple linear mapping on top of both the 1024-dimensional image features and the 312-dimensional attribute vectors to produce a 1,024-dimensional output space. For this dataset we found it helpful to normalize the class prototypes (embedded attribute vectors) to be of unit length, since the attribute vectors come from a different domain than the images. Training episodes were constructed with 50 classes and 10 query images per class. The embeddings were optimized via SGD with Adam at a fixed learning rate of $10^{-4}$ and weight decay of $10^{-5}$. Early stopping on validation loss was used to determine the optimal number of epochs for retraining on the training plus validation set.

\begin{table}[tb]
\caption{Zero-shot classification accuracies on CUB-200.}
\label{tab:cub200}
\begin{center}
\begin{small}
\begin{tabular}{lcc}
\hline
\abovespace
\belowspace
\textbf{Model} &
\begin{tabular}{@{}c@{}}\textbf{Image} \\ \textbf{Features} \end{tabular}  &
\begin{tabular}{@{}c@{}}\textbf{50-way Acc.} \\ 0-shot \end{tabular}  \\
\hline
\abovespace
\textbf{\textsc{ALE}} \cite{akata2013label} & Fisher & 26.9\% \\
\textbf{\textsc{SJE}} \cite{akata2015evaluation} & AlexNet & 40.3\% \\
\textbf{\textsc{Sample Clustering}} \cite{liao2016} & AlexNet & 44.3\% \\
\textbf{\textsc{SJE}} \cite{akata2015evaluation} & GoogLeNet & 50.1\% \\
\textbf{\textsc{DS-SJE}} \cite{reed2016learning} & GoogLeNet & 50.4\% \\
\textbf{\textsc{DA-SJE}} \cite{reed2016learning} & GoogLeNet & 50.9\% \\
\belowspace
\textbf{\textsc{Proto. Nets (Ours)}} & GoogLeNet & \textbf{54.6\%} \\
\hline
\end{tabular}
\end{small}
\end{center}
\end{table}

Table~\ref{tab:cub200} shows that we achieve state-of-the-art results by a large margin when compared to methods utilizing attributes as class meta-data. We compare our method to other embedding approaches, such as ALE \citep{akata2013label}, SJE \citep{akata2015evaluation}, and DS-SJE/DA-SJE \citep{reed2016learning}. We also compare to a recent clustering approach \citep{liao2016} which trains an SVM on a learned feature space obtained by fine-tuning AlexNet \citep{krizhevsky2012imagenet}. These zero-shot classification results demonstrate that our approach is general enough to be applied even when the data points (images) are from a different domain relative to the classes (attributes).

\section{Related Work}
The literature on metric learning is vast \citep{kulis2012metric,bellet2013survey}; we summarize here the work most relevant to our proposed method. Neighborhood Components Analysis (NCA) \citep{goldberger2004neighbourhood} learns a Mahalanobis distance to maximize K-nearest-neighbor's (KNN) leave-one-out accuracy in the transformed space. 
\citet{salakhutdinov2007learning} extend NCA by using a neural network to perform the transformation. Large margin nearest neighbor (LMNN) classification \citep{weinberger2005distance} also attempts to optimize KNN accuracy but does so using a hinge loss that encourages the local neighborhood of a point to contain other points with the same label. The DNet-KNN \citep{min2009deep} is another margin-based method that improves upon LMNN by utilizing a neural network to perform the embedding instead of a simple linear transformation. Of these, our method is most similar to the non-linear extension of NCA \citep{salakhutdinov2007learning} because we use a neural network to perform the embedding and we optimize a softmax based on Euclidean distances in the transformed space, as opposed to a margin loss. A key distinction between our approach and non-linear NCA is that we form a softmax directly over \emph{classes}, rather than individual points, computed from distances to each class's prototype representation. This allows each class to have a concise representation independent of the number of data points and obviates the need to store the entire support set to make predictions.

Our approach is also similar to the nearest class mean approach \citep{mensink2013distance}, where each class is represented by the mean of its examples. 
This approach was developed to rapidly incorporate new classes into a classifier without retraining, however it relies on a linear embedding and was designed to handle the case where the novel classes come with a large number of examples. In contrast, our approach utilizes neural networks to non-linearly embed points and we couple this with episodic training in order to handle the few-shot scenario. \citeauthor{mensink2013distance} attempt to extend their approach to also perform non-linear classification, but they do so by allowing classes to have multiple prototypes. They find these prototypes in a pre-processing step by using $k$-means on the input space and then perform a multi-modal variant of their linear embedding. Prototypical networks, on the other hand, learn a non-linear embedding in an end-to-end manner with no such pre-processing, producing a non-linear classifier that still only requires one prototype per class. In addition, our approach naturally generalizes to other distance functions, particularly Bregman divergences.

Another relevant few-shot learning method is the meta-learning approach proposed in \citet{ravi2017meta}. 
The key insight here is that LSTM dynamics and gradient descent can be written in effectively the same way. An LSTM can then be trained to itself train a model from a given episode, with the performance goal of generalizing well on the query points. Matching networks and prototypical networks can also be seen as forms of meta-learning, in the sense that they produce simple classifiers dynamically from new training episodes; however the core embeddings they rely on are fixed after training. 
The FCE extension to matching nets involves a secondary embedding that depends on the support set. However, in the few-shot scenario the amount of data is so small that a simple inductive bias seems to work well, without the need to learn a custom embedding for each episode.

Prototypical networks are also related to the neural statistician \citep{edwards2017towards} from the generative modeling literature, which extends the variational autoencoder \citep{kingma2013auto,rezende2014stochastic} to learn generative models of datasets rather than individual points. One component of the neural statistician is the ``statistic network'' which summarizes a set of data points into a statistic vector. It does this by encoding each point within a dataset, taking a sample mean, and applying a post-processing network to obtain an approximate posterior over the statistic vector. \citeauthor{edwards2017towards} test their model for one-shot classification on the Omniglot dataset by considering each character to be a separate dataset and making predictions based on the class whose approximate posterior over the statistic vector has minimal KL-divergence from the posterior inferred by the test point. Like the neural statistician, we also produce a summary statistic for each class. However, ours is a discriminative model, as befits our discriminative task of few-shot classification.

With respect to zero-shot learning, the use of embedded meta-data in prototypical networks resembles the method of \cite{lei2015predicting} in that both predict the weights of a linear classifier. The DS-SJE and DA-SJE approach of \cite{reed2016learning} also learns deep multimodal embedding functions for images and class meta-data. Unlike ours, they learn using an empirical risk loss. Neither \cite{lei2015predicting} nor \cite{reed2016learning} uses episodic training, which allows us to help speed up training and regularize the model.

\section{Conclusion}

We have proposed a simple method called prototypical networks for few-shot learning based on the idea that we can represent each class by the mean of its examples in a representation space learned by a neural network. We train these networks to specifically perform well in the few-shot setting by using episodic training. 
The approach is far simpler and more efficient than recent meta-learning approaches, and produces state-of-the-art results even without sophisticated extensions developed for matching networks (although these can be applied to prototypical nets as well). We show how performance can be greatly improved by carefully considering the chosen distance metric, and by modifying the episodic learning procedure. We further demonstrate how to generalize prototypical networks to the zero-shot setting, and achieve state-of-the-art results on the CUB-200 dataset. A natural direction for future work is to utilize Bregman divergences other than squared Euclidean distance, corresponding to class-conditional distributions beyond spherical Gaussians. We conducted preliminary explorations of this, including learning a variance per dimension for each class. This did not lead to any empirical gains, suggesting that the embedding network has enough flexibility on its own without requiring additional fitted parameters per class.  Overall, the simplicity and effectiveness of prototypical networks makes it a promising approach for few-shot learning.

\subsubsection*{Acknowledgements}

We would like to thank Marc Law, Sachin Ravi, Hugo Larochelle, Renjie Liao, and Oriol Vinyals for helpful discussions. This work was supported by the Samsung GRP project and the Canadian Institute for Advanced Research.

{
    \small
    \bibliographystyle{plainnat}
    \bibliography{protonet}
}

\clearpage
\appendix

\section{Additional Omniglot Results}

In Table~\ref{tab:omnicompare} we show test classification accuracy for prototypical networks using Euclidean distance trained with 5, 20, and 60 classes per episode.

\begin{table*}[h]
\caption{Additional classification accuracy results for prototypical networks on Omniglot. Configuration of training episodes is indicated by number of classes per episode (``way''), number of support points per class (``shot'') and number of query points per class (``query''). Classification accuracy was averaged over 1,000 randomly generated episodes from the test set.}
\label{tab:omnicompare}
\begin{center}
\begin{small}
\begin{tabular}{lcccccccc}
\hline
\abovespace
&&\multicolumn{3}{c}{\textbf{Train Episodes}} & \multicolumn{2}{c}{\textbf{5-way Acc.}} & \multicolumn{2}{c}{\textbf{20-way Acc.}} \\\\
\belowspace
\textbf{Model} & \textbf{Dist.} & Shot & Query & Way & 1-shot & 5-shot & 1-shot & 5-shot \\
\hline
\abovespace
\textbf{\textsc{Protonets}}     & Euclid. & 1 & 15 & 5  & 97.4\% & 99.3\% & 92.0\% & 97.8\% \\
\textbf{\textsc{Protonets}}     & Euclid. & 1 & 15 & 20 & 98.7\% & 99.6\% & 95.4\% & 98.8\% \\
\belowspace
\textbf{\textsc{Protonets}}     & Euclid. & 1 & 5  & 60 & 98.8\% & 99.7\% & 96.0\% & 99.0\%\\
\hline
\abovespace
\textbf{\textsc{Protonets}}     & Euclid. & 5 & 15 & 5  & 96.9\% & 99.3\% & 90.7\% & 97.8\% \\
\textbf{\textsc{Protonets}}     & Euclid. & 5 & 15 & 20 & 98.1\% & 99.6\% & 94.1\% & 98.7\% \\
\belowspace
\textbf{\textsc{Protonets}}     & Euclid. & 5 & 5 & 60 & 98.5\% & 99.7\% & 94.7\% & 98.9\% \\
\hline
\end{tabular}
\end{small}
\end{center}
\end{table*}

\begin{figure}[tb]
    \centering
    \includegraphics[width=0.7\linewidth]{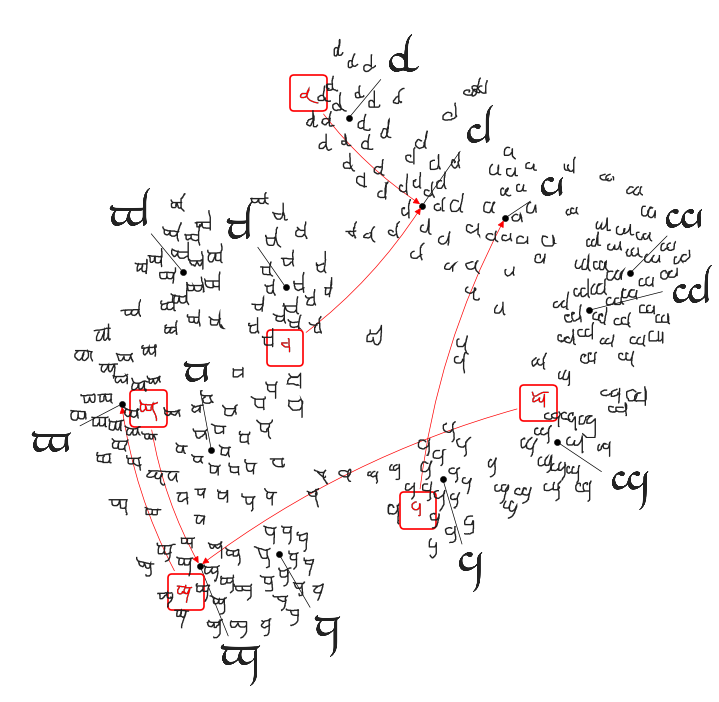}
    \caption{A t-SNE visualization of the embeddings learned by  prototypical networks on the Omniglot dataset. A subset of the Tengwar script is shown (an alphabet in the test set). Class prototypes are indicated in black. Several misclassified characters are highlighted in red along with arrows pointing to the correct prototype.}
    \label{fig:tengwar}
\end{figure}

Figure~\ref{fig:tengwar} shows a sample t-SNE visualization \citep{maaten2008visualizing} of the embeddings learned by prototypical networks. We visualize a subset of test characters from the same alphabet in order to gain better insight, despite the fact that classes in actual test episodes are likely to come from different alphabets. Even though the visualized characters are minor variations of each other, the network is able to cluster the hand-drawn characters closely around the class prototypes. 

\section{Additional \emph{mini}ImageNet Results}

In Table~\ref{tab:miniImagenetcompare} we show the full results for the comparison of training episode configuration in Figure~2 of the main paper. 

\begin{figure*}[bt]
     \centering
     \includegraphics[width=0.9\textwidth]{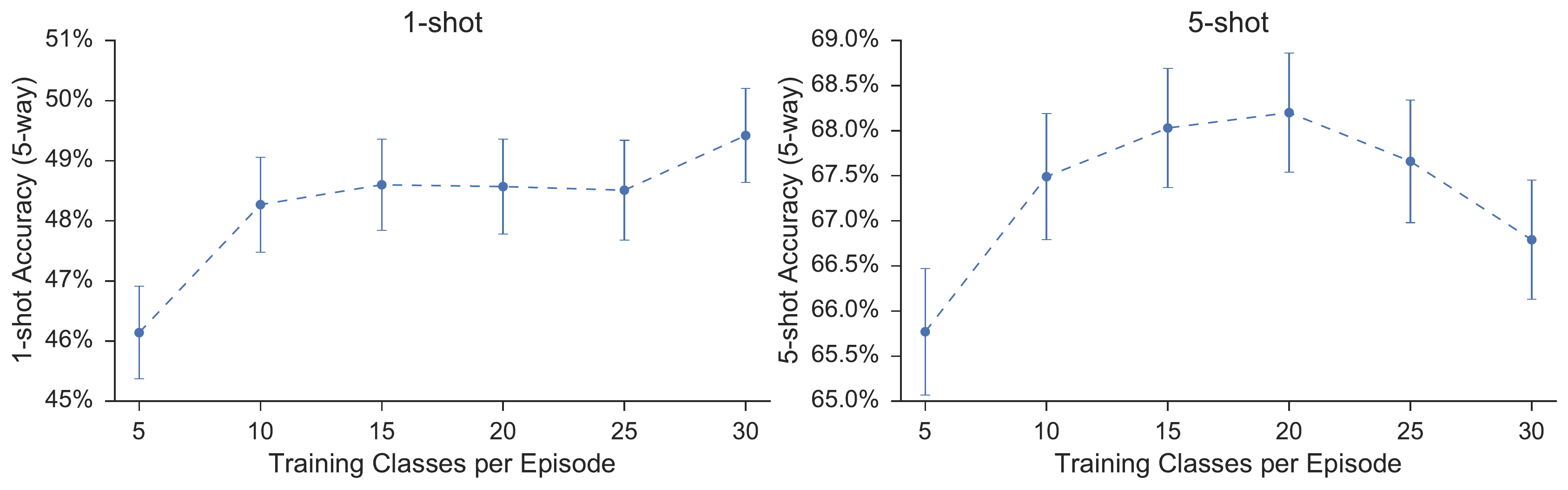}
     \caption{Comparison of the effect of training ``way'' (number of classes per episode) for prototypical networks trained on \emph{mini}ImageNet. Each training episode contains 15 query points per class. Error bars indicate 95\% confidence intervals as computed over 600 test episodes.}
     \label{fig:datacomp}
 \end{figure*}

We also compared Euclidean-distance prototypical networks trained with a different number of classes per episode. Here we vary the classes per training episode from 5 up to 30 while keeping the number of query points per class fixed at 15. The results are shown in Figure~\ref{fig:datacomp}.
Our findings indicate that construction of training episodes is an important consideration in order to achieve good results for few-shot classification.
Table~\ref{tab:miniImagenetway} contains the full results for this set of experiments. 

\begin{table*}[ht]
\caption{Comparison of matching and prototypical networks on \textit{mini}ImageNet under cosine vs. Euclidean distance, 5-way vs. 20-way, and 1-shot vs. 5-shot. All experiments use a shared 
encoder for both support and query points with embedding dimension 1,600 (architecture and training details are provided in Section~3.2 of the main paper). Classification accuracy is averaged over 600 randomly generated episodes from the test set and 95\% confidence intervals are shown.}
\label{tab:miniImagenetcompare}
\begin{center}
\begin{small}
\begin{tabular}{lcccccc}
\hline
\abovespace
&&\multicolumn{3}{c}{\textbf{Train Episodes}} & \multicolumn{2}{c}{\textbf{5-way Acc.}} \\
\belowspace
\textbf{Model} & \textbf{Dist.} & Shot & Query & Way & 1-shot & 5-shot \\
\hline
\abovespace
\textbf{\textsc{Matching Nets / Protonets}} & Cosine  & 1 & 15 & 5 & 38.82 $\pm$ 0.69\% & 44.54 $\pm$ 0.56\% \\
\textbf{\textsc{Matching Nets / Protonets}} & Euclid. & 1 & 15 & 5 & 46.61 $\pm$ 0.78\% & 59.84 $\pm$ 0.64\% \\
\textbf{\textsc{Matching Nets / Protonets}} & Cosine  & 1 & 15 & 20 & 43.63 $\pm$ 0.76\% & 51.34 $\pm$ 0.64\% \\
\belowspace
\textbf{\textsc{Matching Nets / Protonets}} & Euclid. & 1 & 15 & 20 & 49.17 $\pm$ 0.83\% & 62.66 $\pm$ 0.71\% \\

\hline

\abovespace
\textbf{\textsc{Matching Nets}} & Cosine  & 5 & 15 & 5 & 46.43 $\pm$ 0.74\% & 54.60 $\pm$ 0.62\% \\
\textbf{\textsc{Matching Nets}} & Euclid. & 5 & 15 & 5 & 46.43 $\pm$ 0.78\% & 60.97 $\pm$ 0.67\% \\
\textbf{\textsc{Matching Nets}} & Cosine  & 5 & 15 & 20 & 46.46 $\pm$ 0.79\% & 55.77 $\pm$ 0.69\% \\
\textbf{\textsc{Matching Nets}} & Euclid. & 5 & 15 & 20 & 47.99 $\pm$ 0.79\% & 63.66 $\pm$ 0.68\% \\
\textbf{\textsc{Protonets}}     & Cosine  & 5 & 15 & 5 & 42.48 $\pm$ 0.74\% & 51.23 $\pm$ 0.63\% \\
\textbf{\textsc{Protonets}}     & Euclid. & 5 & 15 & 5 & 44.53 $\pm$ 0.76\% & 65.77 $\pm$ 0.70\% \\
\textbf{\textsc{Protonets}}     & Cosine  & 5 & 15 & 20 & 42.45 $\pm$ 0.73\% & 51.48 $\pm$ 0.70\% \\
\belowspace
\textbf{\textsc{Protonets}}     & Euclid. & 5 & 15 & 20 & 43.57 $\pm$ 0.82\% & 68.20 $\pm$ 0.66\% \\
\hline
\end{tabular}
\end{small}
\end{center}
\end{table*}

\begin{table*}[ht]
\caption{Effect of training ``way'' (number of classes per training episode) for prototypical networks with Euclidean distance on \textit{mini}ImageNet. The number of query points per class in training episodes was fixed at 15. Classification accuracy is averaged over 600 randomly generated episodes from the test set and 95\% confidence intervals are shown.}
\label{tab:miniImagenetway}
\begin{center}
\begin{small}
\begin{tabular}{lcccccc}
\hline
\abovespace
&&\multicolumn{3}{c}{\textbf{Train Episodes}} & \multicolumn{2}{c}{\textbf{5-way Acc.}} \\
\belowspace
\textbf{Model} & \textbf{Dist.} & Shot & Query & Way & 1-shot & 5-shot \\
\hline
\abovespace
\textbf{\textsc{Protonets}} & Euclid. & 1 & 15 & 5  & 46.14 $\pm$ 0.77\% & 61.36 $\pm$ 0.68\% \\
\textbf{\textsc{Protonets}} & Euclid. & 1 & 15 & 10 & 48.27 $\pm$ 0.79\% & 64.18 $\pm$ 0.68\% \\
\textbf{\textsc{Protonets}} & Euclid. & 1 & 15 & 15 & 48.60 $\pm$ 0.76\% & 64.62 $\pm$ 0.66\% \\
\textbf{\textsc{Protonets}} & Euclid. & 1 & 15 & 20 & 48.57 $\pm$ 0.79\% & 65.04 $\pm$ 0.69\% \\
\textbf{\textsc{Protonets}} & Euclid. & 1 & 15 & 25 & 48.51 $\pm$ 0.83\% & 64.63 $\pm$ 0.69\% \\
\belowspace
\textbf{\textsc{Protonets}} & Euclid. & 1 & 15 & 30 & 49.42 $\pm$ 0.78\% & 65.38 $\pm$ 0.68\% \\

\hline

\abovespace
\textbf{\textsc{Protonets}} & Euclid. & 5 & 15 & 5  & 44.53 $\pm$ 0.76\% & 65.77 $\pm$ 0.70\% \\
\textbf{\textsc{Protonets}} & Euclid. & 5 & 15 & 10 & 45.09 $\pm$ 0.79\% & 67.49 $\pm$ 0.70\% \\
\textbf{\textsc{Protonets}} & Euclid. & 5 & 15 & 15 & 44.07 $\pm$ 0.80\% & 68.03 $\pm$ 0.66\% \\
\textbf{\textsc{Protonets}} & Euclid. & 5 & 15 & 20 & 43.57 $\pm$ 0.82\% & 68.20 $\pm$ 0.66\% \\
\textbf{\textsc{Protonets}} & Euclid. & 5 & 15 & 25 & 43.32 $\pm$ 0.79\% & 67.66 $\pm$ 0.68\% \\
\belowspace
\textbf{\textsc{Protonets}} & Euclid. & 5 & 15 & 30 & 41.38 $\pm$ 0.81\% & 66.79 $\pm$ 0.66\% \\
\hline
\end{tabular}
\end{small}
\end{center}
\end{table*}

\end{document}